\documentclass{article}

\usepackage{arxiv}

\usepackage[utf8]{inputenc} 
\usepackage[T1]{fontenc}    
\usepackage{hyperref}       
\usepackage{url}            
\usepackage{booktabs}       
\usepackage{amsfonts}       
\usepackage{nicefrac}       
\usepackage{microtype}      
\usepackage{lipsum}		
\usepackage{graphicx}
\usepackage{natbib}
\usepackage{doi}
\usepackage{algorithm}
\usepackage{algorithmic}

\title{3D Point Cloud Completion with Geometric-Aware Adversarial Augmentation}

\author{ Mengxi Wu\\
	New York University\\
	\texttt{mw4355@nyu.edu} \\
	\And
	Hao Huang \\
	New York University\\
	\texttt{hh1811@nyu.edu} \\
	\And
	Yi Fang \\
	New York University\\
	\texttt{yfang@nyu.edu} \\
}

\date{}

\begin{document}
\maketitle

\begin{abstract}
	With the popularity of 3D sensors in self-driving and other robotics applications, extensive research has focused on designing novel neural network architectures for accurate 3D point cloud completion. However, unlike in point cloud classification and reconstruction, the role of adversarial samples in 3D point cloud completion has seldom been explored. In this work, we show that training with adversarial samples can improve the performance of neural networks on 3D point cloud completion tasks. We propose a novel approach to generate adversarial samples that benefit both the performance of clean and adversarial samples. In contrast to the PGD-$k$ attack, our method generates adversarial samples that keep the geometric features in clean samples and contain few outliers. In particular, we use principal directions to constrain the adversarial perturbations for each input point. The gradient components in the mean direction of principal directions are taken as adversarial perturbations. In addition,  we also investigate the effect of using the minimum curvature direction. Besides, we adopt attack strength accumulation and auxiliary Batch Normalization layers method to speed up the training process and alleviate the distribution mismatch between clean and adversarial samples. Experimental results show that training with the adversarial samples crafted by our method effectively enhances the performance of PCN on the ShapeNet dataset.
\end{abstract}

\keywords{Point Cloud \and Shape Completion \and Adversarial Learning \and Geometric Features}

\section{Introduction}
Modern deep learning models are vulnerable to adversarial examples \citep{szegedy2014intriguing}. The existence of adversarial examples is often treated as a threat to the safety-critical application of deep learning models. For example, in point cloud classification or reconstruction, adversarial samples can mislead the model to predict the wrong class or reconstruct different types of objects. Thus, it has triggered a large amount of research focusing on attack/defense techniques in these two fields. However, very few efforts have been made to investigate how the adversarial samples affect the 3D point cloud completion.

In this paper, we investigate adversarial samples and adversarial training in 3D point completion tasks. Xie et al. \citep{Xie_2020_CVPR} demonstrate that training with adversarial samples does help image recognition. Thus, rather than focusing on defending the adversarial samples, we pay attention to use adversarial features to boost the performance of the model on shape completion tasks. However, not all adversarial features are beneficial. Outliers are parts of adversarial features, but they are noise that may distort the original shapes of the point clouds. Training with adversarial samples that have many outliers will let the neural networks learn the noise features instead of geometric properties of the objects that may benefit the performance.  Widely applied attack methods, including FGSM \citep{goodfellow2015explaining} and PGD-$k$ \citep{10.1145/3052973.3053009}, usually generate adversarial samples that contain many outliers. Thus, developing a method to craft strong adversarial samples with few outliers becomes an essential task.

To resolve the outliers issue, various algorithms \citep{Xiang_2019_CVPR, 8803770, liu2020adversarial, yang2021adversarial} have attempted to achieve imperceptible adversarial perturbations. These methods constrain adversarial point clouds to be close to clean ones under specific distance metrics. More recently, in \citep{9294112}, the authors point out that sharp inconsistencies exist between the human perception of 2D images and that of 3D shapes. They suggest that adversarial point clouds should satisfy the surface properties of the clean ones. Technically, they make the magnitudes of local curvatures between adversarial and benign point clouds close to each other. Inspired by their work, we also take the consistency of surface properties into account. We use directions of curvatures on each point to maintain the surface properties. Our approach can be more computation-efficient under some training schemes.

 At each point on the surface, along with different directions, we can obtain different curvatures. The gradient can be decomposed in different curvature directions. We notice that moving a point in the direction of minimum curvature will likely introduce imperceptible changes. However, considering the some special cases, we propose to use gradient components in the mean direction of maximum and minimum curvatures as adversarial perturbations. Our method produces adversarial point clouds without introducing noticeable modifications. We name this attack PMPD-$k$, short for $k$-iteration projection on mean principle attack. Besides, the auxiliary batch normalization method \citep{Xie_2020_CVPR} is applied to migrate distribution mismatch between adversarial and clean samples, and attack strength accumulation \citep{Zheng_2020_CVPR} is applied to reduce time consumption for adversarial samples generation.

To our best knowledge, our work is the first to show that adversarial samples can improve the performance of deep learning models in the 3D point cloud completion task. Compared to the results of vanilla PCN \citep{yuan2018pcn}, the average chamfer distance is reduced by $5.2\%$ when we test PCN trained with PMPD-1 on clean samples in the seen categories. The improvement of robustness is also notable. PMPD-1 helps PCN decrease $4.2\%$ on average chamfer distance when we test the model with adversarial samples in the seen categories. 

\section{Related Work}
\subsection{3D Point Cloud Completion}
Traditional methods adopt voxel grids or distance fields to represent 3D objects \citep{Dai_2017_CVPR, Han_2017_ICCV, Stutz_2018_CVPR}. Based on these representations, convolution neural networks achieve outstanding performances in the tasks of 3D reconstruction \citep{choy20163d, DBLP:journals/corr/GirdharFRG16} and shape completion \citep{Dai_2017_CVPR, Han_2017_ICCV, Zhirong15CVPR}. However, the group of methods based on 3D convolution suffers significant memory consumption and computational cost. Thus, in recent years, researchers have shifted their attention to unstructured point clouds. Compared to voxel grids or distance fields, 3D point cloud representations need less memory consumption and have a strong ability to represent fine-grained details. Point clouds are unordered point sets, so the convolution operation is no longer applicable. For point cloud completion, PCN \citep{yuan2018pcn} is the first learning-based architecture. It contains an encoder that summarizes the geometric information in the input point cloud. A folding-based operation that maps the 2D points onto a 3D surface is adopted in the decoder to approximate a smooth surface representing a shape's local geometry. After PCN, many other methods \citep{Wang_2020_CVPR, Huang_2020_CVPR,Liu_Sheng_Yang_Shao_Hu_2020} are proposed to achieve higher resolution. Most recent work, PoinTr \citep{yu2021pointr} reformulates point cloud completion as a set-to-set translation. It achieves state-of-art results by employing a transformer encoder-decoder architecture.

\subsection{Data Augmentation}
Data augmentation applies a set of transformations to training samples. It is regarded as an essential and effective role to alleviate the overfitting of deep learning models. Traditional data augment methods include translation, random flipping, shifting, scaling, and
rotation. Many
works \citep{wang2019frustum,shi2020pv} present that data augmentation methods improve the performance of neural networks on
3D object detection. Besides these traditional methods, for the 3D point cloud completion task, researchers suggest sampling different partial points clouds from the ground truth during training \citep{yu2021pointr}. Given a center, they remove the furthest $n$ points from the center and take the remaining points as the input partial point cloud. The partial point cloud changes as the center changes. However, training samples generated by this method are always part of the ground truth. No perturbations are added to the coordinates of each point. The arrangement of points remains the same. 

\subsection{PGD-$k$ Attack based Adversarial Training} Adversarial training is an effective approach to
increase the robustness of models against adversarial attacks. In adversarial training, adversarial attacks are taken as a data augmentation method. Models
trained with adversarial samples can achieve considerable robustness.  Adversarial training can be formulated as a min-max optimization problem \citep{madry2018towards}.
\begin{equation}
\min_{f\in \mathcal{H}} \mathbb{E}_{(x,y)\sim \mathcal{D}}[\max_{\delta \in \mathcal{S}} L(f(x+\delta), y)] 
\end{equation}
where $\mathcal{H}$ is the hypothesis space, $\mathcal{D}$ is the distribution of
the training dataset, $L$ is a loss function, $f$ is the neural network, and $\mathcal{S}$ is the allowed perturbation space. A clean sample $x$ will be turned into a perturbed sample by adding adversarial perturbations $\delta$ that maximizes the loss with respect to the ground truth result during the adversarial training process. Models are trained on adversarial samples.

Solving the inner maximization problem is very difficult. PGD-$k$ ($k$-step projected gradient descent \citep{45816}) is an iterative attack algorithm that finds an approximate solution for the inner maximization problem. It takes multi-step projected gradient descent on a negative loss function and projects adversarial perturbations outside the allowed perturbation space $S$ back to $S$:
\begin{equation}
x_{t+1} = \Pi_{x+S}(x_t + \alpha \textrm{sign}(\nabla_{x_t} L(f(x_t), y)))
\end{equation}
In the above equation, $x_t$ is the adversarial sample in the $t$-th
attack iteration and $\alpha$ is the attack step size. $\Pi$ is the projection function that projects adversarial samples back to the
allowed perturbation space $\mathcal{S}$. Our proposed adversarial samples generation method is inspired by the PGD-$k$ attack.

\section{Methodology}
Our geometric-aware adversarial training method includes three major techniques:
\begin{itemize}
\item A novel geometric-aware attack method (PMPD-$k$) is used to generate adversarial samples with very few outliers and maintain the surface properties of clean samples.
\item We reuse adversarial perturbations from the previous epoch to achieve attack strength accumulation across epochs.
\item The model is trained with both adversarial and clean samples. Adversarial samples and clean samples are normalized by different Batch Normalization layers. 
\end{itemize}
 The pipeline of the whole training method is shown in Figure 1, and the details about the algorithm are formally presented in Algorithm 2. The overall idea is as follows. The algorithm first loads the adversarial samples generated from the last epoch and passes them to PMPD-$k$ to compute the adversarial samples in the current epoch. Then, the loss on clean samples is computed with main BN layers, and the loss on adversarial samples is computed with auxiliary BN layers. The parameters of the neural networks are optimized with respect to the sum of two losses. 
 
 \begin{figure*}
\centering
\includegraphics[width=16cm]{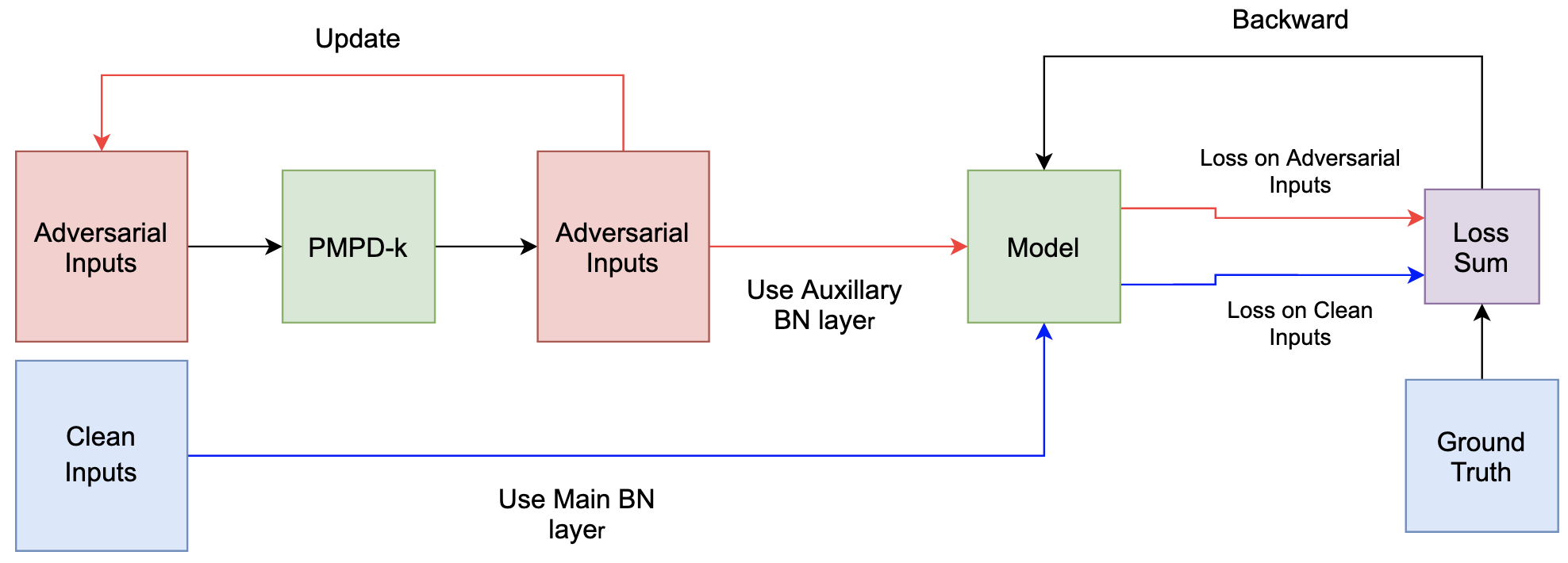}
\caption{The Pipeline of Geometric-Aware Training Algorithm. Initially, adversarial inputs are same as clean inputs. PMPD-$k$ will produce stronger adversarial samples based on the perturbations in the previous epoch.}
\end{figure*}
 
\subsection{PMPD-$k$ Attack}
Our novel attack method, PMPD-$k$ Attack (k-iteration projection on the mean of principal directions attack), is formally presented in Algorithm 1. Its most notable feature is constraining the adversarial perturbations with geometric properties of clean samples. It projects the gradient of an input point on the mean direction of the principal directions at that point and uses the projection as the adversarial perturbations. The adversarial samples generated by this method maintain the geometric features of the clean samples and contain very few outliers.

\subsubsection{Benefits of Principal Directions} We observe that, in the direction of the curvature with minimum absolute value, the surface may bend most slightly compared to other directions. In other words, the surface is closest to the tangent plane in this direction. If we add perturbations to the point along this direction on the tangent plane, the point will still be close to the original surface. It is hard to determine the direction of curvature with minimum absolute value directly. However, since the normal vector will always point toward the side of concavity on a curve, when the surface is upward or downward concave, the curvature with minimum absolute value is equivalent to minimum curvature. The direction of minimum curvature can be easily determined by calculating the principal directions. Real-world 3D objects usually have complicated shapes. Saddle points appear very frequently. Thus, instead of solely using the minimum curvature direction, we use the mean direction of maximum and minimum curvature so that fewer gradient components will be added in the direction of the minimum curvature direction at the saddle points. Figure 2 illustrates our idea.

Using principal directions can be computation-efficient.
Wen et al. \citep{9294112} demonstrate that magnitudes of local curvatures can also effectively describe the bending condition of a surface. However, directly using magnitude differences between adversarial and clean samples as a restriction or regularization term is not very computation-efficient. The magnitudes of the local curvatures need to be recomputed once new perturbations are added to the adversarial samples. For a training scheme that clean samples are fixed during the training process, principal directions of clean samples can be pre-computed and saved before the training process begin. In this way, significant computation consumption can be reduced.

\begin{figure}
\centering
\includegraphics[width=8cm]{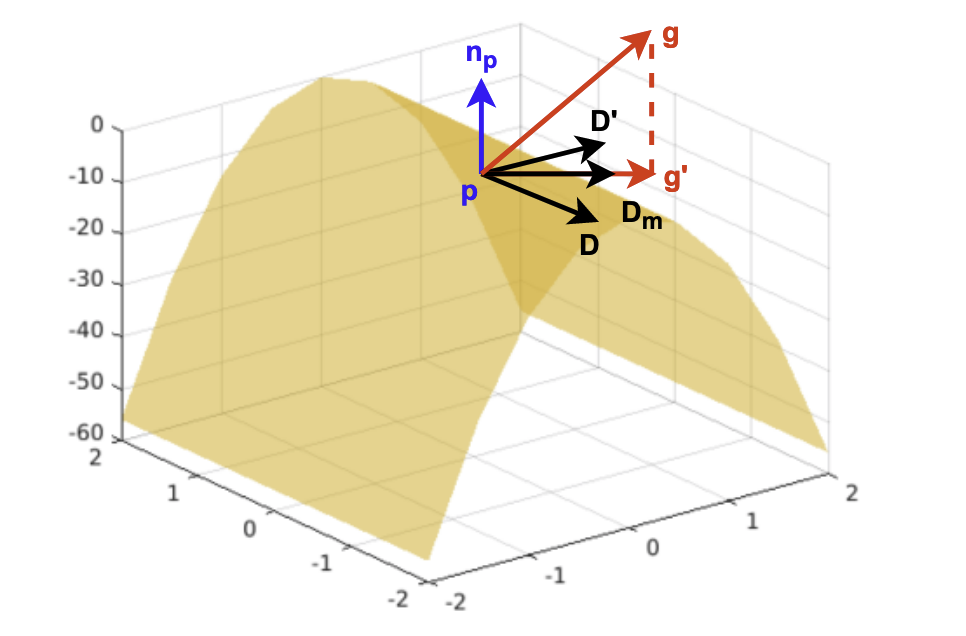}
\caption{PMPD-$k$ Attack. $n_p$ is the normal vector of the tangent plane at point $p$. $D$ and $D'$ are principal directions at point $p$. $D_m$ is the mean direction of minimum and maximum curvature directions. $g$ is the gradient of the $p$. $g'$ is component of $g$ in direction of $D_m$. We use $g'$ as the perturbations for $p$.}
\end{figure}

\begin{algorithm}[tb]
\caption{PMPD-$k$ Attack}
\label{alg:algorithm}
\textbf{Input}: Training sample $x$, Ground truth $gt$, Mean direction of principal directions $d_{m}$\\
\textbf{Parameter}: Movement multiplier $\beta$, Maximum iteration $k$, Network parameter $\theta$, Allowed perturbations space $S$\\
\textbf{Output}: Adversarial sample $x_{adv}$
\begin{algorithmic}[1] 
\STATE Initialize $x_{adv}$ by cloning $x$
\FOR{$i$=1 to $k$}
\STATE Compute loss $l = L(\theta, x_{adv}, gt)$
\STATE Compute gradient $g = \nabla_{x_{adv}} l$
\STATE Compute projection $p =$ rowMulti$(g,d_{m})$ \footnotemark
\STATE Compute perturbation $\delta =$ rowMulti$(p,d_{m})$
\STATE $x_{adv} = \Pi_{x+S}(x_{adv} + \beta \cdot \delta)$
\ENDFOR
\STATE \textbf{return} $x_{adv}$
\end{algorithmic}
\end{algorithm}
\footnotetext{rowMulti denotes row-wise dot product operation.}

\subsubsection{Principal Directions Estimation}
  We adopt the approaches in previous works \citep{6643668,9294112} to compute the estimation of the minimum curvature direction. Let $P$ represents the surface of an input point cloud. For evey point $p \in P$, the unit normal vector $n_p$ of the surface at $p$ is estimated first. We find closest point $p' \in P$ by
$p' =$ argmin$_{q\in P}\|q-p\|_2^2$. We can obtain a local neighbourhood $N_p \subset P$ for each $p$. The number of points in $N_p$ is represented by $k_p$. We use $k_p = 20$. A positive semidefinite covariance matrix $C_p$ can be constructed by
\begin{equation}
C_p = \sum_{q\in N_p}(p-q)\otimes(p-q)
\end{equation}
where $\otimes$ denotes the outer product operation. The estimated $n_p$ is the eigenvector corresponding to the smallest eigenvalue of $C_p$. Thus, we can obtain the estimated $n_p$ through applying eigen-decomposition to $C_p$. 

For every point $p \in P$, we collect the normal vectors of its $k_n$-nearest neighbors. We set $k_n$ to be 10.  This set of normal vectors is represented by $M_p$. Since the principal directions at point $p$ are orthogonal to the normal vector $n_p$, we pick a pair  of  orthogonal  directions in the tangent plane of point $p$ to start. We generate $k_d$ pairs orthogonal directions by rotating this pair of directions on the tangent plane. The $k_d$ we use is 18. All pair of directions are equally spaced. The rotation angle is $\pi / k_n$. For each pair of orthogonal directions ($D_i, D_i')$, we compute the normal variation. In  direction  $D_i$, we find two normal vectors $n_1, n_2 \in M_p$ that have smallest angles with $D_i$. We calculate $d_{12}$, the direction of the intersection line of the plane determined by $n_1$ and $n_2$ and the plane determined by $D_i$ and $n_p$. 
\begin{equation}
d_{12} = (n_1 \otimes n_2) \otimes D_i'
\end{equation}
Then, we use two vectors $n_3,n_4 \in M_p$ that have largest angles with $D_i$ to calculate $d_{34}$ in the same way. The angle between $d_{12}$ and $d_{34}$ is $\alpha_i$, as shown in Figure 3. We repeat the process for $D_i'$ to obtain angle $\beta_i$. The normal variation for $D_i$ and $D_i'$ is $\Delta \theta_i = |\alpha_i - \beta_i|$. The estimated principal directions are the pair of orthogonal directions with largest normal variation. We set the mean direction of minimum and maximum curvature directions to be $D_i + D_i'$. Then, we normalize mean direction by dividing its magnitude.

\begin{figure}
\centering
\includegraphics[width=6cm]{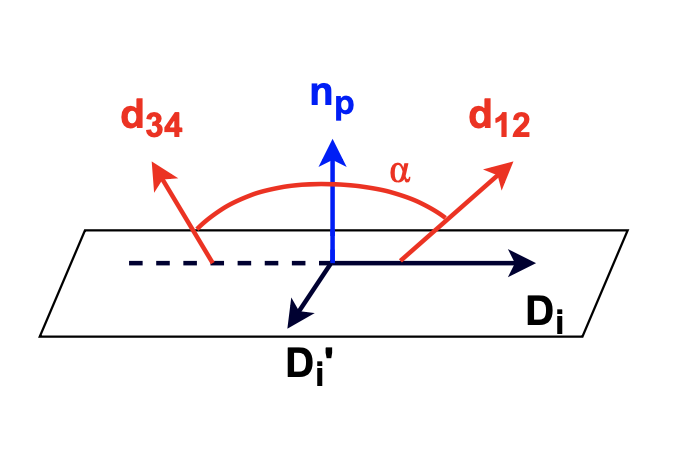}
\caption{Normal Variation Calculation. $n_p$ is the normal vector of the tangent plane. $D_i$ and $D_i'$ are principal directions. $d_{12}$ and $d_{34}$ lie on the plane determined by $n_p$ and $D_i$. $\alpha$ is the angle between $d_{34}$ and $d_{12}$.}
\end{figure}

\subsection{Attack Strength Accumulation}
PGD-$k$ generates adversarial samples with higher loss when the value of $k$ is larger \citep{madry2018towards}. However, with large k, the training process can be very slow. As shown in \citep{Zheng_2020_CVPR},  PGD-$k$ based adversarial training may require close to 100x more training time until the model converges than the natural training process does. Like PGD-$k$, PMPD-$k$ also has a huge time consumption problem.

The key insight of \citep{Zheng_2020_CVPR} is that models from neighboring epochs have high transferability between each other. The adversarial images for the model in the previous epoch will also be the adversarial images for the model in the current epoch. They conclude that the attack strength can be accumulated across epochs. Inspired by their results, we use attack strength accumulation in our training algorithm to let PMPD-$k$ Attack achieve higher attack strength with much fewer attack iterations in each epoch. However, in the early stages of training, model parameters
tend to vary drastically. Hence, the adversarial perturbations in early epochs
will be less useful for subsequent epochs. To reduce the impact of early perturbation, we let the accumulation restart from the beginning periodically.

\subsection{Auxiliary Batch Normalization} To improve performance on clean samples, we train with both clean and adversarial point clouds. However, adding adversarial samples to the clean training set may lead to performance degradation on clean samples. Very limited amount of researches in the point cloud processing community address this issue. Here we refer to a study for image classification. In \citep{Xie_2020_CVPR}, the authors demonstrate that adversarial images and clean images
have different underlying distributions. Maintaining the same Batch Normalization layers for adversarial images and clean images will lead to inaccurate statistics estimation. Inaccurate estimations will cause performance degradation. The distribution mismatch could also be a problem in our case. Therefore, we adapt their auxiliary Batch Normalization method to avoid performance degradation caused by this factor.

\begin{algorithm}[tb]
\caption{Geometric-Aware Adversarial Training}
\label{alg:algorithm}
\textbf{Input}: Clean training set $X$ with ground truth, Mean direction of principal directions $d_{m}$ for clean samples, The number of
epochs to reset perturbation $reset$\\
\textbf{Parameter}: Network parameter $\theta$\\
\textbf{Output}: Network parameter $\theta$
\begin{algorithmic}[1] 
\STATE Initialize $\theta$
\STATE Initialize adversarial training set $X_{adv}$ by cloning $X$
\FOR{$epoch$ = $1,...,N$}
\FOR{each training step}
\STATE Sample a mini-batch $x^b \subset X$, $x_{adv}^b \subset X_{adv}$ with ground truth $gt^b$ and $d_{m}^b$
\IF{$epoch$ $\%$ $reset$ $== 0$}
\STATE$x_{adv}^b = x^b$
\ENDIF
\STATE Compute $x_{adv}^b=$PMPD-$k(\theta, x_{adv}^b, gt^b, d_{m}^b)$ with the auxiliary BNs
\STATE Compute loss $L(\theta, x^b, gt^b)$ with the main BNs
\STATE Compute loss $L(\theta, x_{adv}^b, gt^b)$ with the auxiliary BNs
\STATE $\theta =$ argmin$_{\theta}$ $L(\theta, x^b, gt^b) + L(\theta, x_{adv}^b, gt^b)$
\ENDFOR
\ENDFOR
\STATE \textbf{return} $\theta$
\end{algorithmic}
\end{algorithm}

\section{Experiments} We conduct our test on the PCN model \citep{yuan2018pcn}. We train the PCN with a similar training scheme as described in its original papers as the baseline of our experiments. To show that the direction of perturbations is essential, we also train a PCN model with the widely used PGD-$k$ method and the gradient projection method proposed by Liu et al. \citep{liu2019extending}. This method directly projects the gradients of a point on the tangent plane at that point. Unlike PMPD-$k$, it does not constrain the direction of the gradient components lying on the tangent plane. The gradient components can be in arbitrary directions. We use GD-$k$ to denote this method.
Additionally, we test adding the perturbations solely in the minimum curvature direction. This method we name PMCD-$k$ ($k$-iteration projection on minimum curvature direction attack). Among the above methods, except the training process for baselines, auxiliary batch normalization method and attack strength accumulation method are applied to all other training processes. To speed up the training process, we set the hyper-parameter $k$ to be 1. For brevity, we denote PMDP-1, PMCD-1, GD-1 and PGD-1 as PMDP, PMCD, GD, and PGD.

\subsection{Dataset}
We use the dataset provided by the authors of the PCN model.  They use synthetic CAD models from
ShapeNet to create the PCN dataset. They choose to use a synthetic dataset to generate training data since they can obtain complete, detailed 3D models of objects. The real-world datasets lose a lot of detailed information about the objects. They also found that real-world datasets such as ScanNet \citep{dai2017scannet} or
S3DIS \citep{armeni2017joint} have missing regions due to the limitations
of the scanner’s view. Thus, both are not qualified options for ground truth data. The PCN dataset contains pairs of partial and complete 3D point clouds. There are 30974 models from 8 categories: airplane, cabinet, car, chair, lamp, sofa, table, vessel. The complete point
clouds are 16384 points uniformly sampled on the mesh surfaces. The partial point clouds are generated by back-projecting 2.5D depth images into 3D. The distributions of the partial point clouds created in this way are closer to the real-world sensor data. For every model, eight randomly distributed viewpoints are selected. Eight partial point clouds corresponding to the eight viewpoints are generated, respectively. The size of each partial point cloud is not fixed. Each generated partial point cloud may contain a different number of points. Among the 30974 models, 150 models are reserved for evaluation, and 100 models are reserved for validation. Another test set contains 150 models from 8 novel categories that are not in the training set. Four out of eight categories are similar to the training set, and the rest four categories are dissimilar to the training set.

\subsection{Evaluation Metric} In original papers of PCN and PoinTr, Chamfer Distance is used as evaluation metrics. We follow these two works. Chamfer Distance can measure the average closest point distance between the prediction point cloud and the ground-truth point cloud. Given two point sets $P$ and $P'$, the Chamfer distance between $P$ and $P'$ can be computed as 
\begin{equation}
CD = \frac{1}{n'} \sum_{p'\in P'}\min_{p \in P} \|p-p'\|_2 \\
 +  \frac{1}{n} \sum_{p\in P}\min_{p' \in P'} \|p'-p\|_2
\end{equation}
where $n'$ represents the number of points in $P'$ and $n$ represents the number of points in $P$. To calculate Chamfer Distance, $P$ and $P'$ can have different sizes.

\subsection{Training Schemes}
The optimizer for training PCN models is Adam \citep{kingma2017adam} optimizer. The models are trained for 30  epochs and a batch size of 64. For the baseline, the initial learning rate is 0.001, while for adversarial training, the initial learning rate is set to be 0.0005. The learning rate is decayed by 0.7 every 5 epochs. We reset the accumulation of attack strength every 15 epochs.

\begin{table*}[h]
\caption{Seen Categories}
\centering
\begin{tabular}{l|l|llllllll}
\hline 
Method & Avg & Chair &
Table &
Sofa &
Cabinet &
Lamp &
Car &
Airplane &
Vessel \\
\hline
\hline
 Baseline & 0.01166 & 0.01292 & 0.01135 & 0.01356 & 0.01304  & 0.01472 & 0.00984 & 0.00625 & 0.11201\\
 PGD & 0.01144
 & 0.01298 & 0.01117  &  0.01360 & 0.01206  & 0.01433   & 0.00982 & 0.00655 & 0.01101
\\
GD & 0.01128 & 0.01307 & 0.01080 & 0.01352 & 0.01208 & 0.01368 & 0.00982 & 0.00642 & 0.01089\\
PMCD &0.01110 &
\textbf{0.01253}   & 0.01072   & 0.01353 & 0.01183 &0.01363   & \textbf{0.00957}  &0.00638 &0.01061\\
PMPD & \textbf{0.01105} & 0.01268 & \textbf{0.01054} & \textbf{0.01343} & \textbf{0.01177} & \textbf{0.01360} & 0.00958 & \textbf{0.00616} & \textbf{0.01061}\\
\hline
\end{tabular}
\end{table*}
\begin{table*}[h]
\caption{Seen Categories (with Attack)}
\centering
\begin{tabular}{l|l|llllllll}
\hline 
Method & Avg & Chair &
Table &
Sofa &
Cabinet &
Lamp &
Car &
Airplane &
Vessel 
\\
\hline
\hline
 Baseline & 0.02417 & 0.02619 &
0.02732 &
0.02449  &
0.02585 &
0.02999 &
0.01858 &
0.01905 &
0.02190
\\
 PGD & \textbf{0.02110} &
 \textbf{0.02450} &
\textbf{0.02270} &
\textbf{0.02262} &
\textbf{0.02128} &
\textbf{0.02620} &
\textbf{0.01654} &
\textbf{0.01595} &
\textbf{0.01900}
\\
GD & 0.02266 & 0.02668 &    0.02482   &  0.02510 &   0.02319 &   0.02689  &  0.01774 &   0.01757  &  0.01931\\
PMCD & 0.02355 & 0.02713  &   0.02520    & 0.02523 &    0.02349  & 0.02849  &   0.01809  &  0.01919  &   0.02162\\
PMPD & 0.02316 &
0.02654 & 0.02402  &  0.02544  &  0.02333   & 0.02913  &  0.01852   &   0.01752  &   0.02078
\\
\hline
\end{tabular}
\end{table*}

\begin{table}[h]
\caption{Unseen and Similar Categories}
\centering
\begin{tabular}{l|l|llll}
\hline 
Method & Avg & Bed &
Bench &
Bookshelf & 
Bus\\
\hline
\hline
 Baseline & 0.01611 & \textbf{0.02292} &
0.01267 &
0.01744 &
0.01142 
\\
 PGD & 0.01600 &
 0.02348 & 0.01266 & 0.01642    &   0.01143   
\\
GD & 0.01608 & 0.02406 &  0.01240 &  0.01632 &  0.01153
\\
PMCD & \textbf{0.01590} & 0.02387 & 0.01246  & \textbf{0.01630} &  \textbf{0.01095}
\\
PMPD & 0.01591 & 0.02406 & \textbf{0.01228} & 0.01631 &   0.01098 
\\
\hline
\end{tabular}
\end{table}

\begin{table}[h]
\caption{Unseen and Similar Categories (with Attack)}
\centering
\begin{tabular}{l|l|llll}
\hline 
Method & Avg & Bed &
Bench &
Bookshelf & 
Bus\\
\hline
\hline
 Baseline & 0.03057 & 0.03957 &
0.02852 &
0.03007 &
0.02413 
\\
 PGD & \textbf{0.02607} &
 \textbf{0.03340}   & \textbf{0.02607} &
 \textbf{0.02384}  &       \textbf{0.02096}   
\\
GD & 0.02810 & 0.03738 &    0.02701   &   0.02575 &        0.02227
\\
PMCD & 0.02841 & 0.03727  &  0.02692 &     0.02548  &       0.02395
\\
PMPD & 0.02845 & 0.03767  & 0.02708  &   0.02607    &     0.02296  
\\
\hline
\end{tabular}
\end{table}
\begin{table}
\caption{Unseen and Dissimilar Categories}
\centering
\begin{tabular}{l|l|llll}
\hline 
Method & Avg & Guitar &
Motorbike &
Pistol &
Skateboard\\
\hline
\hline
 Baseline & \textbf{0.01408} & \textbf{0.01242} &
0.01483 &
0.01610 &
0.01298
\\
 PGD & 0.01558 &
 0.01748  & 0.01493   &     \textbf{0.01582} &   0.01410
 
\\
GD & 0.01523 & 0.01547 &  0.01529   &     0.01699  &  0.01318
\\
PMCD & 0.01456 &  0.01424 &  0.01516    &   0.01621   & 0.01262

\\
PMPD & 0.01517 & 0.01629 & \textbf{0.01464}   &     0.01726 &   \textbf{0.01250}

\\
\hline
\end{tabular}
\end{table}
\begin{table}
\caption{Unseen and Dissimilar Categories (with Attack)}
\centering
\begin{tabular}{l|l|llll}
\hline 
Method & Avg & Guitar &
Motorbike &
Pistol &
Skateboard\\
\hline
\hline
 Baseline & 0.02880 & 0.02526 &
0.03250 &
0.03112 &
0.02633 
\\
 PGD & \textbf{0.02606} &
 \textbf{0.02834}   & \textbf{0.02520} &
 \textbf{0.02587}  &       \textbf{0.02486}   
\\
GD & 0.02850 & 0.03011 &    0.02677   &   0.03266 &        0.02445
\\
PMCD & 0.02911 & 0.02573  &  0.03075 &     0.03263  &       0.02736
\\
PMPD & 0.02856& 0.02915 & 0.02783  & 0.03240  &   0.02486  
\\
\hline
\end{tabular}
\end{table}
\subsection{Results}
Figure 4 provides the visualization of completion results of each method on clean samples.  Figure 5 illustrate the visualization results for adversarial samples generated by different methods. We evaluate both performances on clean samples and adversarial samples. The adversarial samples are generated by PGD-5. All results in the six tables are mean chamfer distance per point. Table 1, 3, and 5 are results for clean samples. Table 2, 4, and 6 are results for adversarial samples.

\subsubsection{Performance on Adversarial Samples}
As shown in Table 2, Table 4, all three attack methods perform better than the baseline. In Table 6, PMCD performs slightly worse than the baseline, while PMPD, GD, and PGD still outperform the baseline. PGD performs the best since the adversarial samples generated by it have higher strength. It takes the whole gradients as the adversarial perturbations. The other three methods only take the components of gradient lie on the tangent plane. GD outperforms PMCD and PMPD for a similar reason. Compared with PMCD and PMPD, GD allows the adversarial perturbations have arbitrary direction on the tangent plane so that the input points are perturbed more freely. Thus, the adversarial samples generated by GD may have higher attack strength than the ones generated by PMPD and PMCD.

\subsubsection{Performance on Clean Samples}
From Table 1, baseline performs worst on seen categories among all methods. The other methods are all trained with adversarial samples. This shows that adversarial samples may provide beneficial features to the models and boost the performances. PMPD performs the best. It beats baseline, GD, and PGD on all seen categories. Especially, PMPD outperforms baseline with a considerable decrease in chamfer distance. In addition, PMCD also beats baseline, GD, and PGD in most of the categories. Compared with PMPD and PMCD, GD and PGD lack sophisticated geometric constraints on adversarial perturbations. These results prove the effectiveness of our novel attack method on seen categories.

In Table 3, PMCD has the smallest average value and performs best on bookshelf and bus categories. PMPD performs best on the bench category and has the second smallest average value. Though PMCD and PMPD have smaller gaps with other methods on unseen categories that are similar to the training set compared to seen categories, these two methods still bring the most improvement.

 When tested on unseen categories that are not similar to seen categories, as shown in Table 5, PMPD achieves better performances than the baseline on bench and skateboard categories. However, it does not generalize well on guitar and pistol categories. We compare the baseline with other methods in Table 5, and we found other three methods also do not generalize well on guitar and pistol categories. Thus, we hypothesize that the performance degradation may be caused by the trade-off between robustness and generalization \citep{Raghunathan2019AdversarialTC}. Adversarial samples can improve the robustness of the model but meanwhile can cause performance degradation on examples that are very dissimilar from the training ones. However, PMCD and PMPD still have better results than PGD and GD. This result also shows that constraining the adversarial perturbations with geometric properties is crucial.
 
\begin{figure*}
\centering
\includegraphics[width=16cm]{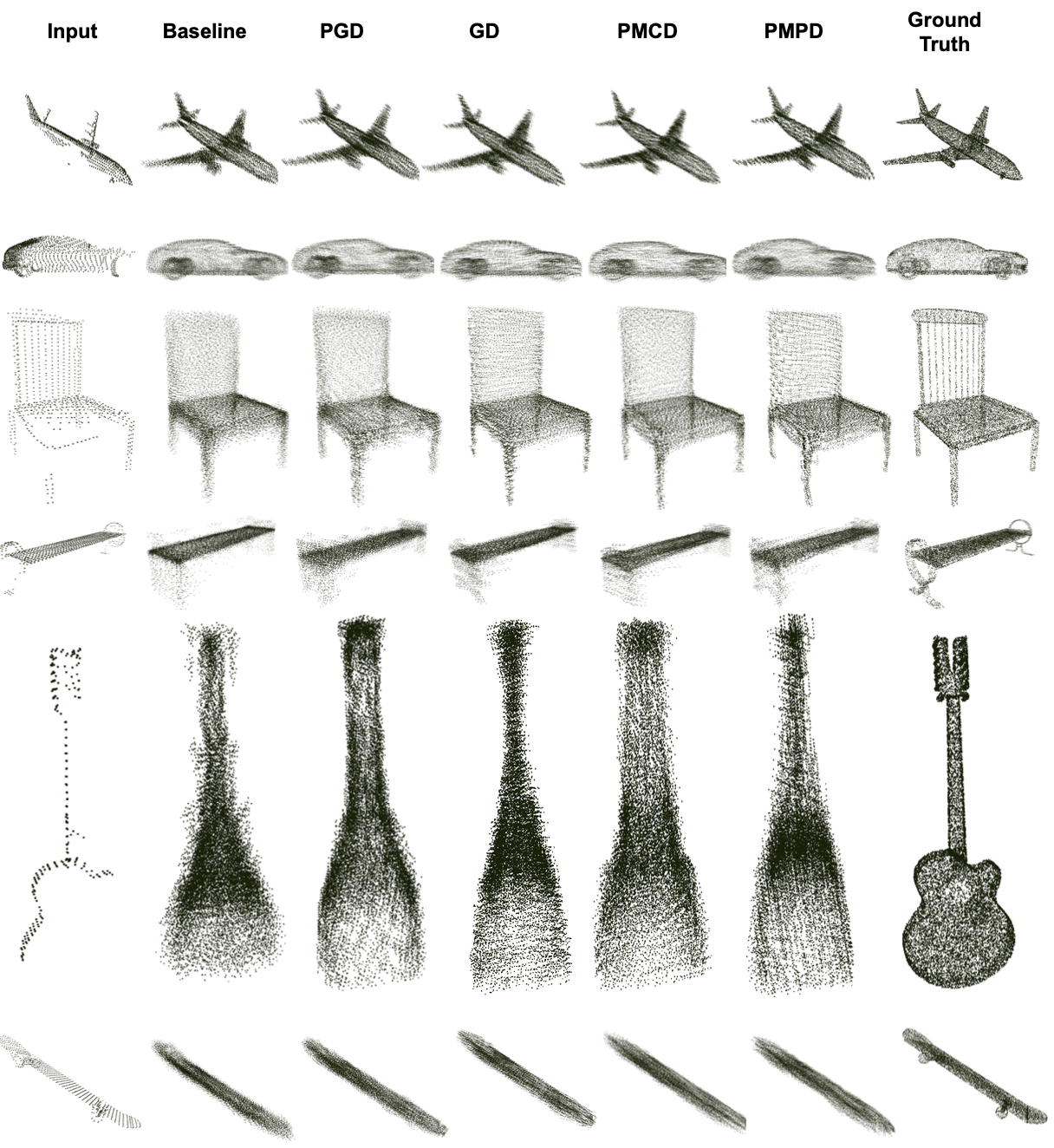}
\caption{Sample Visualization Results of each Method. The first three categories are airplane, car, and chair. These three categories are seen in the training set. The last three categories are bench, guitar, and skateboard. These three categories are not in the training set.}
\end{figure*}

\begin{figure*}
\centering
\includegraphics[width=16cm]{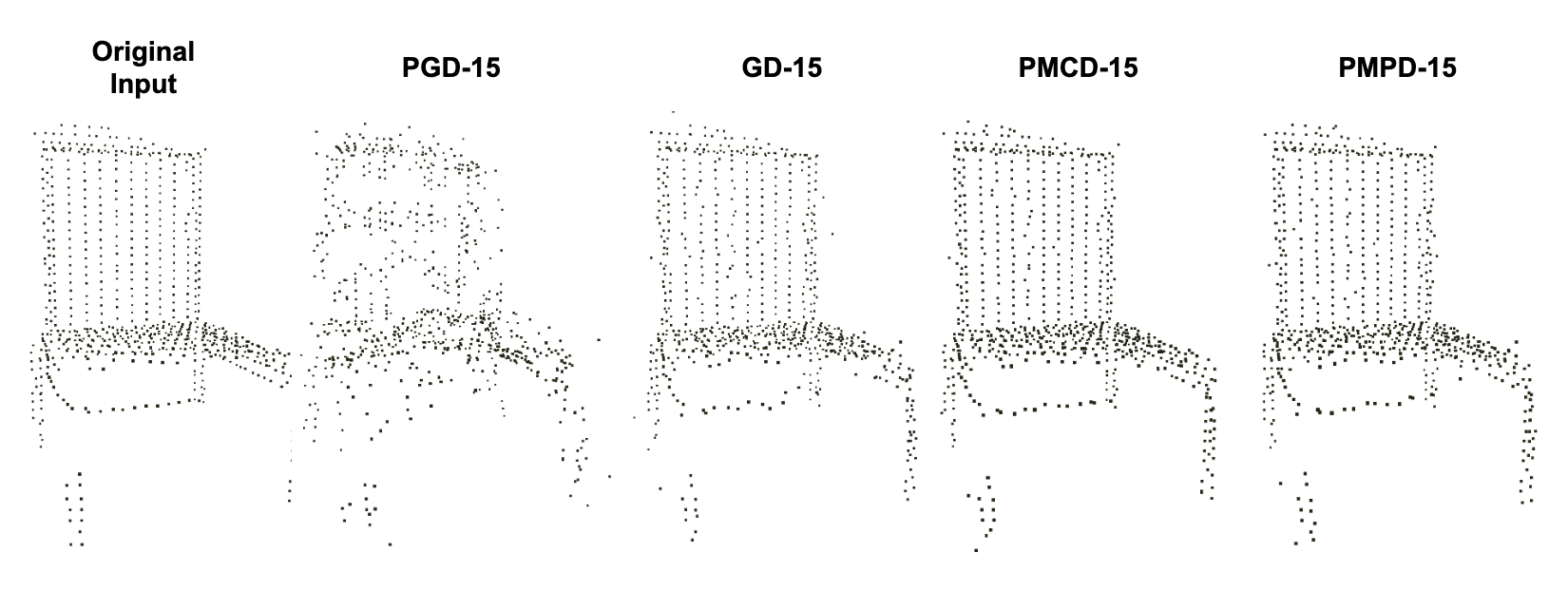}
\caption{Adversarial samples generated by different attack methods. The category we choose is chair. We use 15 iterations to generate adversarial samples. We can see samples generated by PGD or GD contain a lot of outliers, while samples generated by PMCD or PMPD contian very few outliers.}
\end{figure*}

\section{Conclusion}
We introduce a new approach to craft adversarial samples that can benefit deep learning models on shape completion tasks. The key component in our new algorithm is that we constrain the adversarial perturbations by choosing the gradient component in the mean direction of principal directions as the adversarial perturbations. Our approach effectively improves the robustness of the model and the performance of the model on seen categories and unseen categories that are similar to the training samples. For future study, the effect of different weighted combinations of principal directions can be investigated. There may exit a more sophisticated way to mitigate the problems caused by saddle points. Besides, we may further explore methods to reduce the degradation of generalization ability on unseen categories that are dissimilar to the training set.

\bibliographystyle{unsrtnat}
\bibliography{references}

\end{document}